\def\year{2018}\relax
\documentclass[letterpaper]{article} %DO NOT CHANGE THIS
\usepackage{aaai18}  %Required
\usepackage{times}  %Required
\usepackage{helvet}  %Required
\usepackage{courier}  %Required
\usepackage{url}  %Required
\usepackage{graphicx}  %Required
\usepackage[dvipsnames, table]{xcolor}
\usepackage{csquotes}
\usepackage{flushend}
\usepackage{tikz}
\usetikzlibrary{fit,automata,positioning,shapes,shapes.multipart,
arrows,decorations.pathreplacing,matrix} 
\usetikzlibrary{decorations.pathmorphing}
\usetikzlibrary{shapes.callouts} 
\usetikzlibrary{decorations.text}
\usetikzlibrary{patterns} 
\tikzset{text/.default=}
\tikzset{node/.default=}
\usetikzlibrary{matrix}
\usepackage{amssymb}% http://ctan.org/pkg/amssymb
\usepackage{pifont}% http://ctan.org/pkg/pifont
\usepackage{pgfplots}
\usepackage{pgfkeys}
\pgfplotsset{compat=1.3}
\usepackage{xspace}
\newcommand{\mname}{\emph{Pandora}\xspace}

\usepackage{listings} 
\usepackage{enumerate}
\renewcommand{\arraystretch}{1.0}
\usepackage{enumitem}

\newenvironment{my_list}{
   \begin{itemize}
   \setlength{\itemsep}{2.0pt}
   \setlength{\parskip}{-2pt}
   \setlength{\parsep}{-2pt}
}{
   \end{itemize}
}
\usepackage{booktabs}
\usepackage{multirow}
\usepackage{rotating}
\usepackage{makecell}
\usepackage{csvsimple}
\usepackage{pgfplotstable,wrapfig,lipsum}
\begin{document}
\title{Towards Accountable AI: Hybrid Human-Machine Analyses\\ for Characterizing System Failure} 
\author{Besmira Nushi \quad  Ece Kamar \quad  Eric Horvitz
\\
Microsoft Research, Redmond, WA, USA
}
\maketitle
\begin{abstract}
As machine learning systems move from computer-science laboratories into the open world, their accountability becomes a high priority problem. Accountability requires deep understanding of system behavior and its failures. Current evaluation methods such as single-score error metrics and confusion matrices provide aggregate views of system performance that hide important shortcomings. Understanding details about failures is important for identifying pathways for refinement, communicating the reliability of systems in different settings, and for specifying appropriate human oversight and engagement. Characterization of failures and shortcomings is particularly complex for systems composed of multiple machine learned components. For such systems, existing evaluation methods have limited expressiveness in describing and explaining the relationship among input content, the internal states of system components, and final output quality. We present \mname, a set of hybrid human-machine methods and tools for describing and explaining system failures. \mname leverages both human and system-generated observations to summarize conditions of system malfunction with respect to the input content and system architecture. We share results of a case study with a machine learning pipeline for image captioning that show how detailed performance views can be beneficial for analysis and debugging.
\end{abstract}
\section{Introduction}

In light of growing competencies, machine learning and inference are being increasingly pressed into service for analyses, automation, and assistance in the open world. Efforts with machine learning have been undertaken in transportation, healthcare, criminal justice, education, and productivity. Applications include uses of automated perception and classification in high-stakes decisions with significant consequences for people and broader society \cite{dietterich2015rise,DBLP:journals/aim/RussellDT15,DBLP:journals/corr/abs-1709-02435,DBLP:journals/corr/AmodeiOSCSM16}. Standard methods for evaluating the performance of models constructed via machine learning provide a superficial view of system performance. Single-score success metrics, such as the area under the receiver-operator characteristic curve (AUC) or false positive and false negative rates, are computed over test datasets. Such summarizing measures are useful for broad understandings of the statistics of performance and for comparisons across systems. However, they do not provide insights into the details on when and how systems fail. The common metrics on performance are particularly limited for systems composed of multiple components as they fail to reflect how interactions among sets of machine-learned components contribute to errors.

Detailed characterization of system performance through systematic and fine-grained failure analyses has important implications for deploying AI systems in the real world. Identifying and explaining errors at execution time is essential for enabling people to complement or override the system. Understanding when and where errors occur is also a crucial step in detecting and addressing bias. For example, a medical diagnostic system that is 95\% accurate can be experienced as being highly unreliable if the bulk of its errors occur when interpreting symptoms from an under-represented population group. Finally, system designers can use detailed error analyses to make informed decisions on next steps for system improvement.

We present a set of hybrid human-in-the loop and machine learning methods named as \mname that facilitate the process of \emph{describing} and \emph{explaining} failures in machine learning systems. We apply these methods to systems that have a human-interpretable input and that integrate multiple learning components. Failure analysis in these systems presents two main challenges. First, the system may exhibit a \emph{non-uniform} error behavior across various slices of the input space. For instance, the accuracies of a face recognition system may depend on the demographic properties of people (\emph{e.g.,} race, gender, age) \cite{buolamwini2018gender}. Traditional evaluation metrics like error scores and confusion matrices have limited expressiveness in reporting such non-uniformity of error. Second, the complex interactions among the uncertainties of different components can lead to a complex compounding of error that prevents practitioners from understanding the internal dynamics of system failure. 

\begin{figure}[t]
    \centering
      \includegraphics[width=0.7\columnwidth]{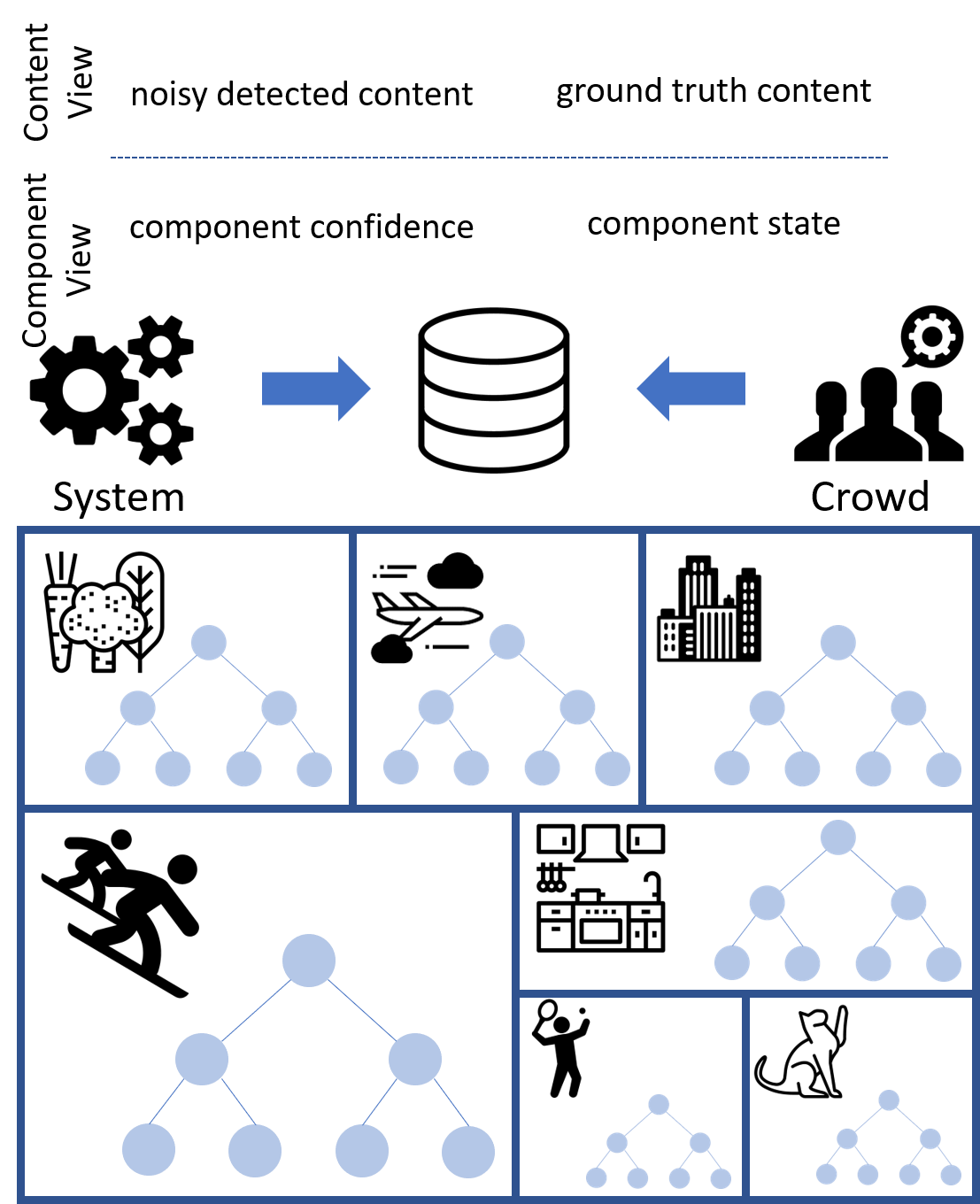}
      \caption{\mname applied to an image captioning system.}
      \label{fig_intro}
  \end{figure}

 \mname addresses these challenges by modeling the relationships among input instances, system execution signals, and errors to develop detailed predictive summaries of performance. First, it clusters the input domain into topical clusters constructed either from human-generated content features or from automated system-generated features representing content. Next, it uses the examples in each cluster to learn interpretable decision-tree classifiers for predicting and summarizing conditions of failure. The classifiers provide transparent \emph{performance views} explaining under which circumstances the system is most likely to err.
 
\mname provides multiple views to highlight different relationships between input data, system execution and system errors. Along one dimension, the views choose the type of data being used for error analysis: signals drawn from content being analyzed or signals collected from component execution. \emph{Content-based views} use detailed ground truth or automatically detected content (\emph{i.e.,} input data) features to learn common situations associated with poor performance. For instance, a face recognizer could report that the system may make more mistakes in recognizing faces of \texttt{old} \texttt{men} wearing \texttt{eyeglasses}. \emph{Component-based views} instead model the relationship between the uncertainty as well as the individual performance state of each component and system failures. For a face recognizer that includes a face detector, component-based views can describe how often the system fails when the detector is uncertain (\emph{i.e.,} low confidence) or wrong (\emph{i.e.,} false detection).
  
  The second dimension for generating views varies the source of data used to provide a multi-faceted error analysis. For this, 
  the views can either choose data generated by the system itself or data collected via human computation. The different views created by varying the data source have complementary purposes: Views generated based on system data characterize performance from an inner system perspective, \emph{i.e.,} what the system knows. Views generated based on crowd input describe system performance from an outer user perspective, \emph{i.e.,} what an ideal system should know (ground-truth). The inner perspective is important to teach the system itself about its failures, whereas the outer perspective helps users understand when to expect failures and cannot be created without the input from human computation. 
The whole process highlights how human computation can be valuable for machine learning not only for simply collecting training data but also for solving complex problems like understanding failure modes.

 Figure~\ref{fig_intro} shows an overview of the approach when applied to an Image Captioning System as a real-world case study. The goal of the system is to automatically generate image captions for given images and it consists of three machine learning components in a pipeline \cite{DBLP:conf/cvpr/FangGISDDGHMPZZ15}. \mname creates content topical clusters (\emph{e.g.,} city, snowboarding, kitchen etc.) and reports the system performance for each distinct cluster. Most importantly, it generates custom decision trees, which uncover surprising facts about the system behavior relevant to its input conditions and the internal state of its architecture. 

 In summary, this work makes the following contributions:
  \begin{my_list}
    \item A new systematic approach for describing and explaining system failure in component-based machine learning systems. \mname supports system designers and machine learning practitioners in understanding system failure and it can be used as an effective tool for debugging and revealing hidden artifacts of system malfunction.
    \item An end-to-end case study of applying \mname to a real-world system for image captioning. The application showcases how such an approach can be implemented in practice to a multi-modal system with a rich input domain as well as multiple machine learning components.
    \item An experimental study for showing and comparing the benefits of the various performance views generated by \mname.
  \end{my_list} 

\section{Background}
\begin{figure} 
\centering
\begin{tikzpicture}
	\footnotesize
	\tikzstyle{text}=[draw=none,fill=none]
	\tikzstyle{box}=[rectangle, draw=MidnightBlue!120]
	
    \node (system) [rectangle,label=below:,fill=lightgray,
    minimum width = 1.0\columnwidth, minimum height = 26 mm]{};
    
    \node[box] (c1) [above left= -2.5 cm
    and -1.8 cm of system, minimum width = 14 mm, minimum height = 10 mm, text width=14 mm,fill=white, very thick,
    align=center] {Visual Detector};    
    
    \node[text] (w1) [above right=1.05cm and -0.2 cm of c1,
    minimum width = 14 mm, text width=17 mm,
    align=center] {\scriptsize{snowboard, 0.96}};
    \node[text] (w2) [below=-0.15 cm of w1,
    minimum width = 14 mm, text width=17 mm,
    align=center] {\scriptsize{ snow (0.94)}};
    \node[text] (w3) [below=-0.15 cm of w2,
    minimum width = 14 mm, text width=17 mm,
    align=center] {\scriptsize{ man (0.89)}};
    \node[text] (w4) [below=-0.15 cm of w3,
    minimum width = 16 mm, text width=19 mm,
    align=center] {\scriptsize{ mountain (0.87)}};
    \node[text] (w5) [below=-0.15 cm of w4,
    minimum width = 14 mm, text width=17 mm,
    align=center] {\scriptsize{ skis (0.71}};
    \node[text] (w6) [below=-0.15 cm of w5,
    minimum width = 14 mm, text width=17 mm,
    align=center] {\scriptsize{$\ldots$}};
    \node[text] (io12) [below=-0.07 cm of w6,
    minimum width = 14 mm, text width=37 mm,
    align=center] {I/O};
    %\node[text] (io12d) [below=0.0 cm of io12,
   % minimum width = 14 mm, text width=37 mm,
    %align=center] {$d_{12}$};
    
    \node[box] (c2) [right=1.55 cm of c1,
    minimum width = 14 mm, minimum height = 10 mm, text width=14 mm,fill=white,very thick,
    align=center]{Language Model};   
    \node[text] (s1) [above right=1.05cm and -1.2 cm of c2,
    minimum width = 14 mm, text width=37 mm,
    align=center] {\scriptsize{A man riding skis}};
    \node[text] (s2) [below=-0.15 cm of s1,
    minimum width = 14 mm, text width=37 mm,
    align=center] {\scriptsize{on a snowy mountain.(-5.62)}};    
    \node[text] (s3) [below=-0.05 cm of s2,
    minimum width = 14 mm, text width=37 mm,
    align=center] {\scriptsize{$\ldots$}};
    \node[text] (s4) [below=-0.05 cm of s3,
    minimum width = 14 mm, text width=37 mm,
    align=center] {\scriptsize{A man flying through}};
    \node[text] (s5) [below=-0.15 cm of s4,
    minimum width = 14 mm, text width=37 mm,
    align=center] {\scriptsize{the air on a snowboard.(-9.51)}};
    \node[text] (io23) [below=0.0 cm of s5,
    minimum width = 14 mm, text width=37 mm,
    align=center] {I/O};
    %\node[text] (io23d) [below=0.0 cm of io23,
    %minimum width = 14 mm, text width=37 mm,
    %align=center] {$d_{23}$};
    
    \node[box] (c3) [right=1.55 cm of c2,
    minimum width = 14 mm, minimum height = 10 mm, text width=14 mm,fill=white,very thick,
    align=center]{Caption Reranker};
    
    \node[inner sep=0pt] (image)  [above= 1.6 cm of c2, draw=gray]
    	{\includegraphics[width=.27\columnwidth]{images/COCO_val2014_000000173245.jpg}};
	
	\draw [draw=Black!100, very thick] (c1) [->] -- (c2);
	\draw [draw=Black!100, very thick] (c2) [->] -- (c3);

    \node[inner sep=0pt] (connleftimage)  [below left= -0.2 and -0.05 of image]
    	{};
    \node[inner sep=0pt] (connleft)  [left= 2.15 of connleftimage]
    	{};
    	
    \node[inner sep=0pt] (connleft2)  [left= 2.1 of connleftimage]
    	{};
    \node[inner sep=0pt] (connleftc1)  [below left=1.6cm and 2.1 of image]
    	{};
	\draw [draw=Black!100, very thick] (connleftimage) [-] -- (connleft);
	\draw [draw=Black!100, very thick] (connleft2) [->] -- (connleftc1);

	\node[inner sep=0pt] (connrightimage)  [below right= -0.2 and -0.05 of image]
    	{};
    \node[inner sep=0pt] (connright)  [right= 2.15 of connrightimage]
    	{};
    	
    \node[inner sep=0pt] (connright2)  [right= 2.1 of connrightimage]
    	{};
    \node[inner sep=0pt] (connrightc1)  [below right=1.6cm and 2.1 of image]
        {};
    \node[inner sep=0pt] (connrightrightc1)  [below right=1.6cm and 2.3 of image]
        {};
    \node[inner sep=0pt] (connrightlabel)  [above =2.4cm  of connrightrightc1]
    	{};
	\draw [draw=Black!100, very thick] (connrightimage) [-] -- (connright);
	\draw [draw=Black!100, very thick] (connright2) [->] -- (connrightc1);
	\node[rectangle, fill=white] (caption) [above=2.3 cm of c3,
    minimum width = 14 mm, minimum height = 10 mm, text width=27 mm,
    align=center]{\#1 \\A man flying through the air on a snowboard. (-4.48)};  
	\draw [draw=Black!100, very thick] (connrightrightc1) [->] -- (connrightlabel);
		
\end{tikzpicture}
\caption{The image captioning system.}
\label{fig:captionsystem}
\end{figure}
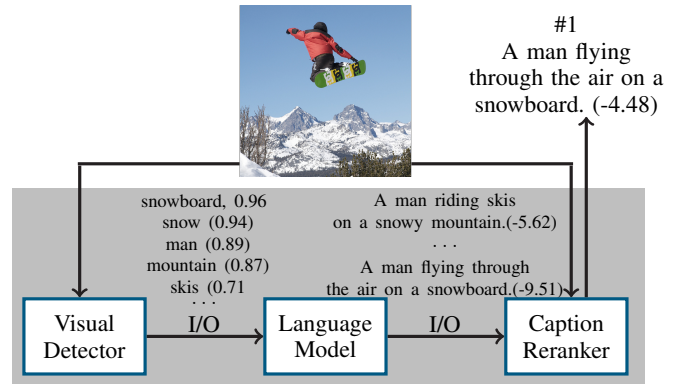
We first describe our main case study, an image captioning system for generating textual descriptions of images. This system was chosen as one of the winners of the 2015 captioning challenge\footnote{http://cocodataset.org/\#captions-challenge2015} and is the blueprint for several applications including the provision of descriptions for assisting visually impaired people\footnote{https://www.microsoft.com/en-us/seeing-ai/}\cite{salisbury2017toward}. Then, we use this system as a running example to illustrate the functionalities of \mname. Figure~\ref{fig:captionsystem} shows the system architecture, consisting of
three machine learning components. The first and third components use
convolutional neural networks% combined with multiple instance learning \cite{zhang2006multiple}. The 
, while the second one is a maximum-entropy language
model \cite{berger1996maximum}.

\noindent{\bfseries{Visual Detector.}} The first component takes an image as an
input and detects a list of words associated with recognition scores. 
%On average, it outputs approximately 25 words per image. 
The detector recognizes only a restricted vocabulary of the 1000 most common words 
in the training captions. The vocabulary contains different parts of speech (\emph{e.g.} verbs, nouns etc.) possibly relevant to the image.

\noindent{\bfseries{Language Model.}} 
This component is a statistical model trained on the human captions for the training images. It generates likely word sequences as captions, based on the words recognized from the Visual Detector, without having access to the input image. The set of the 500 most likely image captions and the respective log-likelihood scores are forwarded to the Caption Reranker. Notice that since the Language Model does not see the image, the log-likelihood captures the linguistic likelihood of the sequence but does not have a notion of how adequate the sentence is for the image.

\begin{table*}[ht]
    \centering
    
    \footnotesize
    \def\arraystretch{1.1}
    \setlength{\tabcolsep}{0.3em}
    \begin{tabular}{c|c|c|}
    \cline{2-3}
    \multicolumn{1}{c|}{}                                                                                                       & \multicolumn{1}{c|}{\begin{tabular}[c]{@{}c@{}}{\bf Crowd Data}\\ (What should an ideal system know?)\end{tabular}} & \multicolumn{1}{c|}{\begin{tabular}[c]{@{}c@{}}{\bf System Data}\\ (What does the system know?)\end{tabular}} \\ \hline
    \multicolumn{1}{|c|}{\multirow{2}{*}{\begin{tabular}[c]{@{}c@{}}{\bf Content View}\\ (When does the system fail?)\end{tabular}}}  &              ground truth objects list                                                                                           &               detected objects list                                                                                            \\
    \multicolumn{1}{|c|}{}                                                                                                      &           ground truth activities list                                                                                             &              detected activities list                                                                                            \\ 
    \multicolumn{1}{|c|}{}                                                                                                      &                       ground truth count \textsubscript{\sc{objects}}                                                                                 &                  detected count \textsubscript{\sc{objects}}                                                                                       \\
    \multicolumn{1}{|c|}{}                                                                                                      &                       ground truth count \textsubscript{\sc{activities}}                                                                                  &                detected count \textsubscript{\sc{activities}}                                                                                        \\
    \cline{1-3}
    
    \multicolumn{1}{|c|}{\multirow{2}{*}{\begin{tabular}[c]{@{}c@{}}
        {\bf Component View}\\(How does the system fail?)
    \end{tabular}}}  
                          & \texttt{precision},\texttt{recall} (objects  \textsubscript{\sc{Visual Detector}}) & \texttt{avg},\texttt{std},\texttt{max},\texttt{min}(confidence \textsubscript{\sc{Visual Detector}}) \\
    \multicolumn{1}{|c|}{}& \texttt{precision},\texttt{recall} (activities  \textsubscript{\sc{Visual Detector}}) & \texttt{avg},\texttt{std},\texttt{max},\texttt{min}(confidence \textsubscript{\sc{Language Model}}) \\ 
    \multicolumn{1}{|c|}{}& \texttt{language},\texttt{commonsense} (sentences  \textsubscript{\sc{Language Model}})  & \texttt{avg},\texttt{std},\texttt{max},\texttt{min}(confidence \textsubscript{\sc{Caption Reranker}}) \\
    \multicolumn{1}{|c|}{}& \texttt{satisfaction} (top10\_captions  \textsubscript{\sc{Caption Reranker}})  & confidence \textsubscript{\sc{Best Caption $\rightarrow$ Visual Detector}} \\
    \multicolumn{1}{|c|}{}& & confidence \textsubscript{\sc{Best Caption $\rightarrow$ Language Model}} \\
    \hline
    \end{tabular}
    \caption{Performance views generated by \mname for the image captioning system and the respective data used as predictive features. Crowd data refers to features that can be acquired only with human input, and system data refers to features that can be extracted from the internal data flowing in the system.}
    \label{tb_views}
\end{table*}
\noindent{\bfseries{Caption Reranker.}}  
The task of the component is to rerank the captions generated from the
Language Model and to select the best match for the image. The
Reranker uses multiple features including the similarity between the
vector representations of both images and captions, the log-likelihood of the sentence from the Language Model, and the length of the sequence. The caption with the highest ranking score is selected as the final best caption.

\noindent{\bfseries{Dataset.}} All components are individually trained on the
MSCOCO dataset \cite{DBLP:conf/eccv/LinMBHPRDZ14}, a case library of captioned photos, constructed to define the image captioning challenge. The database contains 160,000 images as well as five human-generated
captions for all images. We use images randomly sampled from the validation part of the dataset to evaluate
our approach.

\section{Pandora}

\subsection{Problem}
We seek to develop tools that provide more transparent views on performance for system designers. The tools should help designers to understand when and how machine-learned classifiers make mistakes. The approach assumes that the system is composed by a set of modules or components and that the inputs and outputs of the components and the system are interpretable by people. Interpretability facilitates analyses of content via human computation. For interpretable, component-based systems, performance views serve as analytical summaries of system quality that answer the following questions:

\noindent {\bf Q1}: \emph{When does the system fail?} - For a system with a rich input domain or a highly dimensional feature space, evaluation metrics like accuracy, AUC or multi-class confusion matrices do not depict a complete view of system performance. Often, relying only on these generic metrics can hide serious underlying quality issues, especially for data partitions with low representation. Hence, recognizing the input characteristics that are more prominently related to failure, is essential to engineering accountable systems.

\noindent {\bf Q2}: \emph{How does the system fail?} - When a system is composed or makes use of different machine learning services, errors can originate from different parts of the system. It is also not uncommon that failure is caused as a combinatorial state of multiple smaller and hidden errors in the composing components. Although blame assignment is challenging for such systems \cite{sculley2014machine,DBLP:conf/aaai/NushiKHK17}, it is nevertheless important to know how internal component failures and uncertainties are related to the overall system failure. 

\subsection{Performance views}
\mname uses two types of data to answer the above questions. Content-based views use content features to describe \emph{when} the system fails, while component-based views use component state features to explain in depth \emph{how} the system fails. Content and component feature data can be collected from two different data sources: human feedback in the form of crowdsourcing tasks (ground-truth data) or internal system data flows (noisy system signals). Table \ref{tb_views} shows in a grid style a summary of the two dimensions (data type and data source) along with the features being used from each view for the image captioning system. Below, we explain in detail the meaning of these features. In the last part of the section, we also briefly describe how this process would look for another multi-modal interactive system.
%face recognition system. 

\noindent {\bf Content-based views} create a mapping between the input and the overall system failure. For the image captioning system, each input instance is represented by a list of objects and activities. For views created from crowdsourcing data, this list consists of words (verbs or nouns) that crowdsourcing workers think should be mentioned in the final caption. This gives a ground truth perspective to the view and helps system designers to understand the types of input content that the system handles well (or not). Such perspective can also be helpful for reporting performance to customers and end users. 

For views created from system data, the list of objects and activities is the list of words recognized from the Visual Detector as shown in Figure~\ref{fig:captionsystem}. Since automatic detection is not always correct, this perspective provides information on how the system's self-knowledge relates to its output state. For example, by using this analysis, one can identify detections that are commonly linked to system failure. 

\noindent {\bf Component-based views} 
model how internal component dynamics lead to errors. Depending on the data source, features here can express: i) the component output quality as judged by crowdsourcing workers, or ii) the component output confidence as perceived by the component itself. The component output quality is evaluated as follows. For every image in the evaluation dataset, we expose the output of each component to workers and ask them to correct the output, taking into consideration the task of the component and the system as a whole. This feedback simulates the performance of the system with ground-truth output for components, which can then be used to compute quality performance metrics. For instance, workers can flag sentences generated by the Language Model that have bad \texttt{language} or no \texttt{commonsense} (see Table~\ref{tb_views}). The number of such sentences per image can then be used as a predictive and descriptive feature for output failure. The views built using output quality data show the sensitivity of the system to internal errors.% For example, for our case study, this view revealed that the system is more sensitive to low precision in the Visual Detector rather than low recall (See Results section).

The component view that models the relationship between component confidence scores and system errors makes use of a number of handcrafted features, computed as follows: For every image, we aggregate the confidence of the component as the \texttt{average}, \texttt{standard deviation}, \texttt{maximum}, and the \texttt{minimum} of all output scores. In particular, the \texttt{maximum}(confidence \textsubscript{\sc{Caption Reranker}}) feature represents the final confidence score of the Caption Reranker for the best caption. Besides the individual component scores, the feature set also includes scores linked to the best caption with respect to previous components in the pipeline. For example, if the final caption for an image is "A man flying
through the air on a snowboard", we aggregate the Visual Detector confidence scores for the composing words (man, flying, air, snowboard). These features are reflected in features confidence \textsubscript{\sc{Best Caption $\rightarrow$ Visual Detector}}. Similarly, the feature confidence \textsubscript{\sc{Best Caption $\rightarrow$ Language Model}} contains the log-likelihood of the best caption according to the Language Model.

\subsection{View creation and reports}
Now, we describe the process of creating performance views in \mname and the specific reports associated with each view. The end-to-end process is focused on a sample evaluation dataset chosen by the system designer. The sample might come from the training or test dataset depending on the use case, depending on the goal of explaining the current system behavior on the training data versus debugging errors arising when challenged with a previously unseen test set. \mname can create generic as well as clustered reports for a given performance view. Generic reports analyze the evaluation dataset as a whole, while clustered reports decompose the analysis according to distinct semantic clusters. As we show in the experimental evaluation, although there is value in extracting generic failure information, clustered reports are more predictive for system performance and they discover cluster-specific errors, which cannot be identified via generic views.

View generation is a two-step process: 1) clustering the evaluation dataset based on content signals, and 2) detailed reporting globally and per cluster. The process generalizes to all views created in \mname. 

\noindent {\bf Evaluation dataset clustering.} To decompose performance evaluation, \mname initially clusters the evaluation dataset into topical clusters. Each image is represented as a document of objects. Depending on which data source the system designer wants to use (crowd data or system data), the list of objects comes either from the object terms reported by crowdsourcing workers or recognized by the Visual Detector. Note that this list is necessary for any type of clustered performance views (both content and component-based) before generating any report.

We use agglomerative hierarchical clustering \cite{jain1988algorithms} and the Euclidean distance as a similarity measure. While other clustering algorithms may also be relevant, the hierarchical clustering representation brings important usability advantages as it gives freedom to users to reduce the number of clusters by easily merging together similar clusters whose performance they want to explain and evaluate jointly. For example, the crowd data clusters in our analysis contain both \texttt{snowboarding} and \texttt{skiing} separately, but it may also be useful to merge and analyze them together as they share the same parent in the cluster hierarchy.

\noindent {\bf Reports.} The content of failure analysis reports (generic and clustered) consists of:

\noindent 1- \emph{Evaluation metrics} are computed from human satisfaction provided by crowd workers. Clustered variants aggregate satisfaction over selected clusters, highlighting the topical strengths and the weaknesses of the system. 

\noindent 2- \emph{Decision-tree performance predictors} are trained as generic or separate decision tree classifiers per cluster using the features part of the performance view. The choice of decision trees was motivated by the interpretability properties of these classifiers. 
%Since the main purpose of the predictive summary is to convey to system designers conditions of failure, it is essential that the decision making is human interpretable. 
For example, a branch of a decision tree from the component-based views for the image captioning system may summarize failure prediction as: \emph{"If the precision of the Visual Detector is less then 0.8 and there are less than five satisfactory captions for humans in the Caption Reranker, the system fails in 95\% of the cases."}. Moreover, system designers can zoom into the decision tree and explore the concrete instances classified in each leaf. This functionality makes decision trees for performance prediction a tool for fine-grained debugging, error reproducibility, and, perhaps most importantly, a tool for humans to decide when to complement and override the system. 

Finally, it is important to note that such decision tree classifiers are not unique. As the process of training decision trees tends to find a minimal set of features as splitting conditions, the tree frequently omits features that are correlated with the parents. For more extensive explorations, the system designer can intentionally leave a feature out of the tree to investigate other failure conditions in depth or generate feature rankings, as described next. 

\noindent 3- \emph{Feature rankings} identify the most informative features in the view for predicting system failure. We compute the mutual information between the feature and system performance (\emph{i.e.} human satisfaction) as the ranking criterion. The same criterion is used for splitting nodes in the decision trees. Mutual information not only captures the correlation between two variables but also other statistical dependencies that can be useful for failure prediction. Knowing the most important features is beneficial to having a full view of all the failure conditions in the system.
%The experimental evaluation shows examples of concrete reports for the case study and the most important findings.

\subsection{Applications to other systems}
\mname is designed for component-based machine learning systems with human-interpretable input and output for components and for the performance of the system as a whole. We use image captioning as a case study. However, the methodology generalizes to other composable machine learning systems. As an example, consider a multimodal assistant that helps users through single-turn dialog interactions. This hypothetical system combines computer vision for recognizing users as well as speech recognition and dialog systems to understand and communicate with users. The input data is the combination of face images with sound files for human speech. The output is the system response to the human user. This system satisfies the conditions above; the input instances with images and speech are human interpretable as humans can analyze whether the image contains a human face, or transcribe speech. The output of each component (face recognizer, speech recognizer and dialog manager) is human interpretable. Finally, humans can also evaluate the overall system behavior for satisfaction. 

Once these properties are verified, adapting \mname to a new system requires customizing the data collection steps to the system at hand. These customization steps include designing human computation tasks for analyzing the input data content (labeling images and transcribing sound), for correcting the output of each component (face recognition, speech recognition and dialog manager) and for evaluation (analyzing final user satisfaction). In addition to the details of acquiring human effort, customization may also include generating additional features describing system execution to enhance the features given in Table 1 on demand.  The analytical process of \mname would operate in the same way, but on revised features (as per Table 1) developed for the new system to adequately summarize the content, component execution, and final satisfaction.

\section{Crowdsourced Data Collection}
In addition to the data flowing between the system components, \mname makes significant use of data from crowdsourced micro-tasks. These tasks are used for three purposes: 1) system evaluation (for all views), 2) content data (for clustering and for all views using crowd assessments of  ground-truth), and 3) component quality features (for component-based views with crowd data). For these purposes, we leverage the task design and dataset we collected in prior work for troubleshooting the image captioning system \cite{DBLP:conf/aaai/NushiKHK17}. The dataset had been collected mainly for simulating component fixes with crowdsourcing. We adapted the fix data by generating features that directly express the ground truth input content and the original quality of each component.

\noindent {\bfseries{System evaluation.}} There are multiple evaluation metrics for image captioning, including BLEU(1-4) \cite{papineni2002bleu}, CIDEr \cite{vedantam2015cider} and METEOR \cite{banerjee2005meteor}. These metrics are largely adopted from the machine translation domain. They compare the automatic caption to five image captions retrieved from crowdsourcing workers that are made available as part of the MSCOCO dataset. While this evaluation is generally less expensive than directly asking people to report their satisfaction, it does not always correlate well with human satisfaction \cite{anderson2016spice}. Hence, we center this study around direct assessments of human satisfaction with the system output. For this task, we show workers an (image, caption) pair and ask them whether they find the caption satisfactory \"if the caption was used to describe the image to a visually impaired user\".

\noindent {\bfseries{Content data collection.}} As we describe each input with a list of recognized words, the task for ground-truth content data collection corresponds to the task of evaluating the Visual Detector. Here, workers correct the list of objects and activities generated by the Visual Detector by adding or removing items from the list. Workers are instructed to include in the list only words relevant to be mentioned in the caption. The resulting list after majority vote aggregation is considered as the input content for the image. The content is used for i) building ground truth clusters from objects, and for ii) generating content feature data where the presence of each word constitutes a separate binary feature.  

\noindent {\bfseries{Component quality feature data collection.}} 
Component views with crowd data work on the same clustering as the content views. They build features that express the quality of components. For the dataset on fixes acquired in prior work \cite{DBLP:conf/aaai/NushiKHK17}, crowd workers introduced repairs to the component outputs. For the current study, we leverage these fixes to evaluate the original quality state of the components. For instance, by contrasting the output of the Visual Detector before and after the crowd intervention, we can measure both its precision and recall. Similarly, by knowing how many sentences are not commonsense or with bad language from the Language Model, it is possible to find out the percentage of good (or bad) sentences generated by the component. Finally, for the Caption Reranker, workers are allowed to select up to three good captions from the top 10 captions. Based on these signals, we compute features that tell us whether (and how often) the best Reranker caption is selected from workers and whether workers have found any caption in the top 10 set that could be satisfactory. %Both these are included as separate features for the view.

\noindent {\bfseries{Quality control.}} For all crowdsourcing tasks we applied the following techniques for quality control: 1) worker training via examples and online feedback, 2) low-quality work detection based on worker disagreement, and 3) small batching to avoid worker exhaustion and bias reinforcement. 

% \noindent 1- \emph{Worker training} - We paid special attention to the task setup by reiterating over the user interface design, providing multiple examples of correct solutions directly on the interface, and giving online feedback to workers on their work quality.

% \noindent 2- \emph{Spam detection} - Due to the multimodal nature of the case study, the set of crowdsourcing tasks is rich and versatile both in terms of content  (\emph{e.g.} images, words, sentences) and design (\emph{e.g.} yes / no questions, word list correction etc.). Therefore, each task required specific spam detection techniques. Given the lack of ground truth, we used worker disagreement as the main criterion to distinguish low-quality work. However, due to potential subjectivity in our tasks, we used worker disagreement only as an initial indication of low quality and further analyzed distinct cases to decide upon work acceptance.

% \noindent 3- \emph{Small batching} - Large crowdsourcing batches are exposed to low-quality work risk as workers may get tired or reinforce repetitive and wrong biases over time.
%     To avoid such effects, we performed experiments in small-sized batches (\emph{e.g.} 250 image-caption pairs) which were published in periodical intervals. This allowed
%     us to analyze the batch data offline and give constructive feedback to specific workers on how they can improve their work. 
\section{Experimental Evaluation}
We now report on the main findings obtained with applying our approach to the image captioning system. The study is based on an \texttt{evaluation} dataset of 1000 images selected randomly from the validation dataset in MSCOCO \cite{DBLP:conf/eccv/LinMBHPRDZ14}. The evaluation reports that we present here are concrete examples of reports that can be generated with \mname for various performance views. 
\begin{table}[t]
    \centering
    \footnotesize
    \begin{filecontents*}{data/sys_eval_gt.csv}
description,top5,satisfaction
baseball,baseball:field:bat:game:ball,\cellcolor{green!15}0.800
kitchen,kitchen:counter:cabinets:stove:oven,\cellcolor{red!15}0.586
waterfront,water:ocean:beach:man:boat,\cellcolor{red!15}0.604
snowboard,snow:snowboard:slope:hill:mountain,\cellcolor{green!15}0.750
man,man:phone:cell:tie:hat,\cellcolor{red!15}0.455
tennis,tennis:court:racket:player:game,0.656
trains,train:tracks:station:platform:engine,\cellcolor{green!15}0.774
city scene,street:city:traffic:road:people,0.667
vegetables,vegetables:broccoli:food:plate:table,\cellcolor{green!15}1.000
desk,desk:keyboard:laptop:table:computer,0.724
\cellcolor{gray!15}{\bfseries all clusters},\cellcolor{gray!15},\cellcolor{gray!15}{\bfseries 0.578}
\end{filecontents*}
\pgfplotstabletypeset[
        col sep=comma,
        string type,
        every head row/.style={%
    %         before row={\Xhline{1pt}
    %             \multicolumn{5}{l}{Human Satisfaction Scores}  \\ \hline
    %         },
            before row=\Xhline{1pt},
            after row=\Xhline{1pt}
        },
        every last row/.style={after row=\hline},    
        columns/description/.style={column name=Description, column type={p{1.5cm}|}},
        columns/top5/.style={column name=Top 5 objects, column type={p{4.6cm}|}},    
        columns/satisfaction/.style={column name=Satisfactory, column type={p{1.5cm}}}
        ]{data/sys_eval_gt.csv} 
    \caption{System evaluation - Crowd data content clusters.}
    \label{tb:sys_eval_gt}
\end{table}
\begin{table}[t]
    \centering
    \footnotesize
    \begin{filecontents*}{data/performance_gt.csv}
description,human_agreement,content_view,component_view
baseball,0.860,0.900,0.850
kitchen,\cellcolor{red!15}0.807,0.767,0.700
waterfront,0.871,0.670,0.750
snowboard,0.863,0.950,0.850
man,0.867,0.675,0.842
tennis,0.863,0.825,0.775
trains,0.884,0.775,0.833
city scene,\cellcolor{red!15}0.813,0.710,0.700
vegetables,0.832,1.000,1.000
desk,\cellcolor{red!15}0.786,0.867,0.733
\cellcolor{gray!15}{\bfseries all clusters},\cellcolor{gray!15}{\bfseries 0.841},\cellcolor{gray!15}{\bfseries 0.756},\cellcolor{gray!15}{\bfseries 0.747}
\cellcolor{gray!15}{\bfseries generic model},\cellcolor{gray!15},\cellcolor{gray!15}{\bfseries 0.597},\cellcolor{gray!15}{\bfseries 0.711}
\end{filecontents*}
\pgfplotstabletypeset[
        col sep=comma,
        string type,
        every head row/.style={%
    %         before row={\Xhline{1pt}
    %             \multicolumn{5}{l}{Human Satisfaction Scores}  \\ \hline
    %         },
            before row=\Xhline{1pt},
            after row=\Xhline{1pt}
        },
        every last row/.style={after row=\hline},    
        columns/description/.style={column name=Cluster, column type={p{1.9cm}|}},
        columns/human_agreement/.style={column name=Human agreement, column type={p{1.5cm}|}},    
        columns/content_view/.style={column name=Content view, column type={p{1.5cm}|}},
        columns/component_view/.style={column name=Component view, column type={p{1.5cm}}}
        ]{data/performance_gt.csv} 
    \caption{Failure prediction - Crowd data content clusters.}
    \label{tb:performance_gt}
\end{table}
\begin{table}[t]
    \centering
    \footnotesize
    \begin{filecontents*}{data/sys_eval_nongt.csv}
description,top5,satisfaction
baseball,baseball:player:man:ball:field,0.628
kitchen,sink:kitchen:stove:refrigerator:cabinets,\cellcolor{red!15}0.609
tennis,court:man:player:tennis:ball,\cellcolor{red!15}0.523
vegetables,plate:table:vegetables:broccoli:food,\cellcolor{green!15}0.889
animals,dog:cat:man:bear:bed,\cellcolor{red!15}0.321
animals,giraffe:zebra:field:giraffes:zebras,\cellcolor{green!15}0.800
trains,train:tracks:street:man:bus,0.613
skateboard,man:skateboard:skate:trick:person,0.700
surfing,surfboard:board:wave:water:man,\cellcolor{green!15}0.900
desk,computer:table:desk:laptop:keyboard,\cellcolor{red!15}0.556
\cellcolor{gray!15}{\bfseries all clusters},\cellcolor{gray!15},\cellcolor{gray!15}{\bfseries 0.578}
\end{filecontents*}
\pgfplotstabletypeset[
        col sep=comma,
        string type,
        every head row/.style={%
    %         before row={\Xhline{1pt}
    %             \multicolumn{5}{l}{Human Satisfaction Scores}  \\ \hline
    %         },
            before row=\Xhline{1pt},
            after row=\Xhline{1pt}
        },
        every last row/.style={after row=\hline},    
        columns/description/.style={column name=Description, column type={p{1.5cm}|}},
        columns/top5/.style={column name=Top 5 objects, column type={p{4.6cm}|}},    
        columns/satisfaction/.style={column name=Satisfactory, column type={p{1.5cm}}}
        ]{data/sys_eval_nongt.csv} 
    \caption{System evaluation - System data content clusters.}
    \label{tb:sys_eval_nongt}
\end{table}
\begin{table}[t]
    \centering
    \footnotesize
    \begin{filecontents*}{data/performance_nongt.csv}
description,human_agreement,content_view,component_view
baseball,0.851,0.890,0.790
kitchen,\cellcolor{red!15}0.817,0.683,0.783
tennis,0.864,0.770,0.790
vegetables,0.852,0.950,0.967
animals,0.845,0.813,0.777
animals,0.890,0.800,0.925
trains,0.907,0.850,1.000
skateboard,0.870,0.950,0.750
surfing,0.940,0.900,0.900
desk,\cellcolor{red!15}0.775,0.650,0.800
\cellcolor{gray!15}{\bfseries all clusters},\cellcolor{gray!15}{\bfseries 0.841},\cellcolor{gray!15}{\bfseries 0.786},\cellcolor{gray!15}{\bfseries 0.780}
\cellcolor{gray!15}{\bfseries generic model},\cellcolor{gray!15},\cellcolor{gray!15}{\bfseries 0.628},\cellcolor{gray!15}{\bfseries 0.678}
\end{filecontents*}
\pgfplotstabletypeset[
        col sep=comma,
        string type,
        every head row/.style={%
    %         before row={\Xhline{1pt}
    %             \multicolumn{5}{l}{Human Satisfaction Scores}  \\ \hline
    %         },
            before row=\Xhline{1pt},
            after row=\Xhline{1pt}
        },
        every last row/.style={after row=\hline},    
        columns/description/.style={column name=Cluster, column type={p{1.9cm}|}},
        columns/human_agreement/.style={column name=Human agreement, column type={p{1.5cm}|}},    
        columns/content_view/.style={column name=Content view, column type={p{1.5cm}|}},
        columns/component_view/.style={column name=Component view, column type={p{1.5cm}}}
        ]{data/performance_nongt.csv} 
    \caption{Failure prediction - System data content clusters.}
    \label{tb:performance_nongt}
\end{table}

\subsection{System evaluation}
As part of system evaluation, Tables~\ref{tb:sys_eval_gt} and \ref{tb:sys_eval_nongt} summarize clusters that were discovered with ground-truth crowd data and system data respectively, and report system performance for each cluster. Each row shows the top five most frequent words for the cluster and the fraction of instances for which crowd workers found the image caption satisfactory. Due to space restrictions, we present results for 10 (out of 30) representative clusters. %A full report for all clusters will be published in a longer technical report. 

\noindent {\bfseries{Result 1:}} System performance is non-uniform and varies significantly across topical clusters. The tables highlight with green clusters where the system has a high human satisfaction rate ($\geq$ 0.75) and with red for those with low human satisfaction rate ($\leq$ 0.65). For example, we see that the captioning system has a much better performance in crowd data clusters talking about \texttt{baseball} than in clusters about \texttt{kitchen}. Decomposing system performance in this form, provides system designers with insights about the topical strengths and weaknesses of their system.

\noindent {\bfseries{Result 2:}} Reports generated for crowd and system data clusters reveal different insights about the system. By contrasting results in both tables, we see that the \texttt{baseball} cluster constructed by crowd annotations has a higher satisfactory rate than the same cluster built with annotations detected by the Visual Detector: 0.8 versus 0.628. This demonstrates that although the system does well for images that indeed contain baseball, it performs poorly for images where the Visual Detector recognizes words related to baseball. This observation provides hints that the Visual Detector has a high recall but low precision for words relevant to the baseball topic. %A similar situation occurs for clusters like \texttt{trains} and \texttt{desk}. 

\subsection{Performance prediction}
Next, we evaluate the performance prediction accuracy of decision trees generated by \mname, as shown in Table~\ref{tb:performance_gt} and~\ref{tb:performance_nongt}. The accuracy here expresses the fraction of instances for which the decision tree can predict whether the final image caption is going to be satisfactory to crowd workers. Since this is a highly subjective task and people may have slightly different expectations, we include human agreement as a point of reference, \emph{i.e.,} how many workers agree with the majority vote aggregation. The tables highlight in red clusters with the lowest agreement. The main observation with respect to human agreement is that workers have low agreement for images that contain a high number of possibly relevant objects (\emph{e.g.,} \texttt{kitchen}, \texttt{desk}). 
%This is the case for clusters like \texttt{kitchen}, \texttt{desk}, and \texttt{city scene}. Images in these topics usually have multiple objects (\emph{e.g.} for \texttt{kitchen}: sink, fridge, stove, oven, table) and workers often disagree on the amount of required detail.

The prediction accuracy per cluster is averaged over five folds of cross validation within the cluster data. The last two rows of each table show i) the overall accuracy of decision trees for each view per cluster (\emph{i.e.,} all clusters), and ii) the accuracy of views if they would use a single generic decision tree trained on non-clustered data (\emph{i.e.,} generic model). 

\noindent {\bfseries Result 3:} Cluster-specific models are more accurate in predicting system performance due to the non-uniform distribution of errors and as we show in the next reports, the feature set most predictive for failure is also cluster specific. The discrepancy between generic models and cluster models is lower for component views. This shows that condition rules expressing either the components' quality or their confidence generalize better than rules about content.

\noindent {\bfseries Result 4:} Content views and component views are complementary to each other in terms of describing and predicting failure. Although, we do not observe overall differences in the prediction accuracy, for particular clusters one type of view can be better than the other as they describe different failure conditions. For instance, for images where it is crucial to mention specific terms (\emph{e.g.,} \texttt{snowboard}, \texttt{tennis}), content views are more accurate.% For such images, omitting one single important word from the caption results in evident quality drop.

\noindent {\bfseries Result 5:} Performance views trained on system data are at least as accurate as views trained from crowd data. This confirms that features expressing the internal operations of the system can indeed be informative. Although they provide observational evidence versus ground-truth assessments, they contain useful information regarding system confusion. This result highlights the promise of building systems that can predict system failure in real time. %without need for crowd involvement. 

\begin{table}[t]
    \centering
    \footnotesize
    \begin{filecontents*}{data/content_features.csv}
single_crowd,cluster_crowd,single_system,cluster_system
cluster \textsubscript{\sc{obj.}},\cellcolor{gray!20}count \textsubscript{\sc{obj.}},cluster \textsubscript{\sc{obj.}},people
\cellcolor{gray!20}cluster \textsubscript{\sc{act.}},\cellcolor{gray!20}cluster \textsubscript{\sc{act.}},cat,racquet
vegetables,grass,cluster \textsubscript{\sc{act.}},\cellcolor{gray!20}count \textsubscript{\sc{obj.}}
\cellcolor{gray!20}count \textsubscript{\sc{obj.}},people,\cellcolor{gray!20}sitting,swinging
riding,net,bench,grass
zoo,swinging,man,riding
sitting,man,\cellcolor{gray!20}count \textsubscript{\sc{obj.}},women
broccoli,holding,bear,skate
snowboarder,crowd,dog,sign
rice,swing,broccoli,\cellcolor{gray!20}sitting
\end{filecontents*}
\pgfplotstabletypeset[
        col sep=comma,
        string type,
        every head row/.style={%
    %         before row={\Xhline{1pt}
    %             \multicolumn{5}{l}{Human Satisfaction Scores}  \\ \hline
    %         },
            before row=\Xhline{1pt},
            after row=\Xhline{1pt}
        },
        every last row/.style={after row=\hline},    
        columns/single_crowd/.style={column name=Generic model Crowd data, column type={p{1.9cm}|}},    
        columns/cluster_crowd/.style={column name=\texttt{tennis} Crowd data, column type={p{1.5cm}|}},
        columns/single_system/.style={column name=Generic model System data, column type={p{1.9cm}|}},
        columns/cluster_system/.style={column name=\texttt{tennis} System data, column type={p{1.7cm}}}
        ]{data/content_features.csv} 
    \caption{Feature rankings for content views built from a single model and cluster-specific models (\texttt{tennis}).}
    \label{tb:content_features}
\end{table}

\subsection{Feature rankings}
We also make available feature rankings according to their mutual information with human satisfaction. %We use mutual information as a ranking criterion since decision trees presented above use the same criterion for selecting the most important splitting feature conditions. 
We note that not all the highly ranked features are present in the final decision tree. This is because the training algorithm adaptively selects as next splitting conditions those features that are most informative after the parent feature has been observed. Thus, feature rankings are still necessary for completely understanding failure conditions.
\begin{table}[t]
    \centering
    \footnotesize
    \begin{filecontents*}{data/component_features_crowd.csv}
single_crowd
\texttt{satisfaction}(top10\_captions \textsubscript{\sc{Caption Reranker}})
\texttt{precision}(objects \textsubscript{\sc{Visual Detector}})
cluster \textsubscript{\sc{objects}}
\texttt{recall}(objects \textsubscript{\sc{Visual Detector}})
\texttt{commonsense}(sentences \textsubscript{\sc{Language Model}})
\end{filecontents*}
\pgfplotstabletypeset[
        col sep=comma,
        string type,
        every head row/.style={%
    %         before row={\Xhline{1pt}
    %             \multicolumn{5}{l}{Human Satisfaction Scores}  \\ \hline
    %         },
            before row=\Xhline{1pt},
            after row=\Xhline{1pt}
        },
        every last row/.style={after row=\hline},    
        columns/single_crowd/.style={column name=Generic model --- Crowd data, column type={p{6.5cm}}}    
        ]{data/component_features_crowd.csv} 
    \vspace{2pt}
    \begin{filecontents*}{data/component_features_system.csv}
single_system
 \texttt{max}(confidence \textsubscript{\sc{Best Caption $\rightarrow$ Visual Detector}})
 \texttt{max}(confidence \textsubscript{\sc{Visual Detector}})
 \texttt{avg}(confidence \textsubscript{\sc{Visual Detector}})
cluster \textsubscript{\sc{objects}}
 \texttt{avg}(confidence \textsubscript{\sc{Best Caption $\rightarrow$ Visual Detector}})
\end{filecontents*}
\pgfplotstabletypeset[
            col sep=comma,
            string type,
            every head row/.style={%
        %         before row={\Xhline{1pt}
        %             \multicolumn{5}{l}{Human Satisfaction Scores}  \\ \hline
        %         },
                before row=\Xhline{1pt},
                after row=\Xhline{1pt}
            },
            every last row/.style={after row=\hline},    
            columns/single_system/.style={column name=Generic model --- System data, column type={p{6.5cm}}}    
            ]{data/component_features_system.csv} 
    \caption{Feature rankings for component views.}
    \label{tb:component_features}
\end{table}

Table~\ref{tb:content_features} shows the 10 best features extracted from content data for the \texttt{tennis} cluster and for the whole \texttt{evaluation} dataset (generic model). The table highlights in gray the features observed in both generic and cluster models.

\noindent {\bfseries{Result 6:}} The sets of best features for single models and cluster-specific models have low intersection. Features in the intersection are generic features (\emph{e.g.,} clusterings and counts. However, as we show in our previous results, these features are not sufficient for making instance-based decisions that hold for specialized clusters. 

\noindent {\bfseries{Result 7:}} Best content features from crowd data can identify terms that indicate whether the system will perform either very well or very poorly. For example, the system has good satisfactory rate for images that contain \texttt{broccoli}. Best features from system-generated content data have a similar function but they are automatically detected terms. From our observations, many of these features are misrecognitions. A prominent example is the feature \texttt{cat}. The Visual Detector detects a cat for 20\% of the images, while only 3.8\% of the images contain a cat. %Interestingly, because of this explanation, the best content view decision tree generated from system data (as a single model) uses only the \texttt{cat} feature to discriminate between successful and non-successful instances. 
This result shows that some system execution signals (\emph{e.g.,}, \texttt{cat} detection) can identify system confusion and these signals can be discovered by \mname.

Table~\ref{tb:component_features} shows the best features for component views from both crowd and system-generated data. By comparing these sets with those for particular clusters, we note that the intersection is larger for component views. Nevertheless, we notice visible differences in the exact rankings and in the way how these features are used to construct splitting conditions for the decision trees.

\noindent {\bfseries{Result 8:}} Component data views reveal concrete facts about system failure such as the respective importance of precision and recall of the Visual Detector, and the confidence of the Visual Detector on the terms included in the final caption.

\subsection{Decision tree examples}

To illustrate the failure conditions learned by decision trees in \mname, we show tree examples for different types of views. The leaf nodes represent a set of (image,caption) instances, which the decision tree classifies as of the same performance (\emph{i.e.,} green for satisfactory captions and red otherwise). The images displayed below leaf nodes are representative examples for the set. The \emph{samples} tuple shows the number of instances with non-satisfactory captions in its first element and the number of instances with satisfactory captions in the second. The ratio between the two defines how discriminative a leaf node is. This information is available to system designers who can interactively explore failure leaves.

%Not all leaf nodes are equally discriminative. 
%For instance, a leaf node with samples=[7,0] is more discriminative than another one with samples=[6,4]. 

\begin{figure}[t]
    \centering
      \includegraphics[width=1.0\columnwidth]{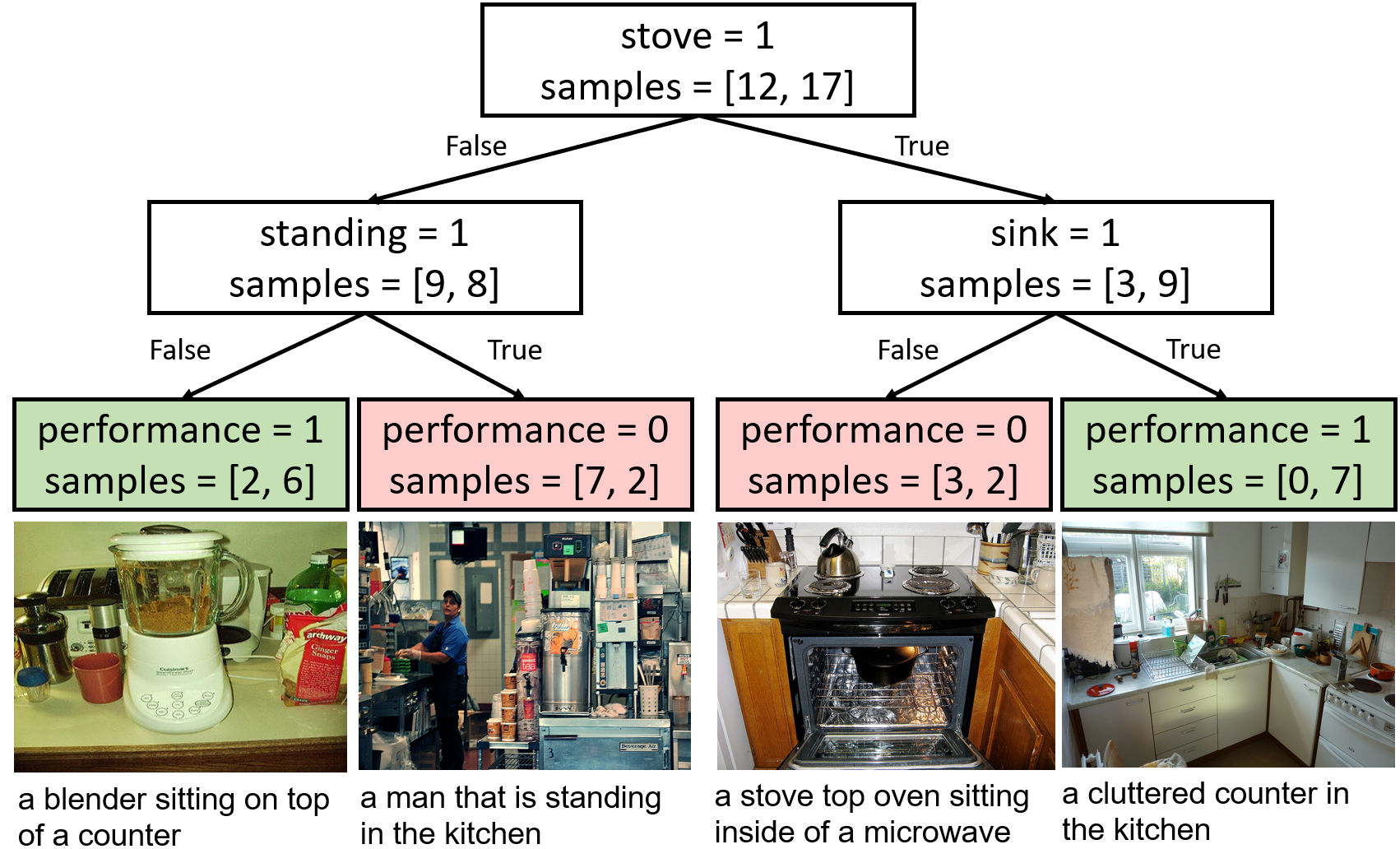}
      \caption{Content view with crowd data for \texttt{kitchen}.}
      \label{fig:kitchen_content_gt}
\end{figure}

Figure~\ref{fig:kitchen_content_gt} visualizes the decision tree for a content view from crowd data for the cluster \texttt{kitchen}. An example of a condition failure that this tree expresses is that the system has a high failure rate for images where someone is standing. After investigating the images in this leaf and comparing outputs with human-generated captions, we find that the standing activity is too generic and that people rate more highly captions referring to more specific kitchen-related activities (\emph{e.g.,} cook, prepare).

\begin{figure}[t]
    \centering
      \includegraphics[width=0.85\columnwidth]{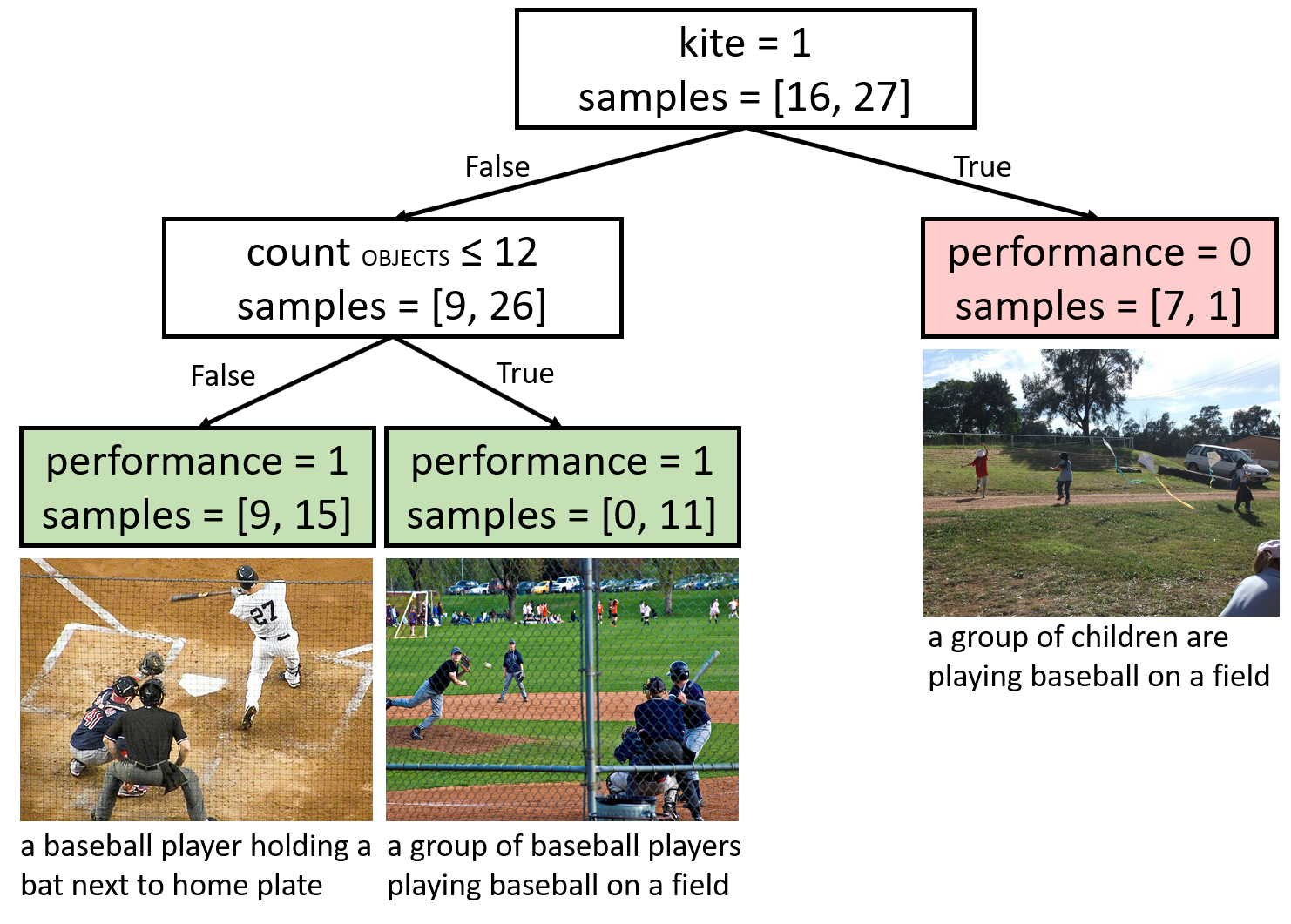}
      \caption{Content view with system data for \texttt{baseball}.}
      \label{fig:baseball_content_nongt}
  \end{figure}

Figure~\ref{fig:baseball_content_nongt} shows a content view built with system data for the cluster \texttt{baseball}. Interestingly, from analyzing the right tree branch (kite=1), one can notice that these are images for which the Visual Detector recognizes both \texttt{baseball} and \texttt{kite} scenes. In fact, the caption is wrong most of the time when the Detector recognizes a kite in this cluster. The system on the other hand performs very well for images in which it does not detect a kite and it detects less than 12 objects. 

\begin{figure}[t]
    \centering
      \includegraphics[width=1.0\columnwidth]{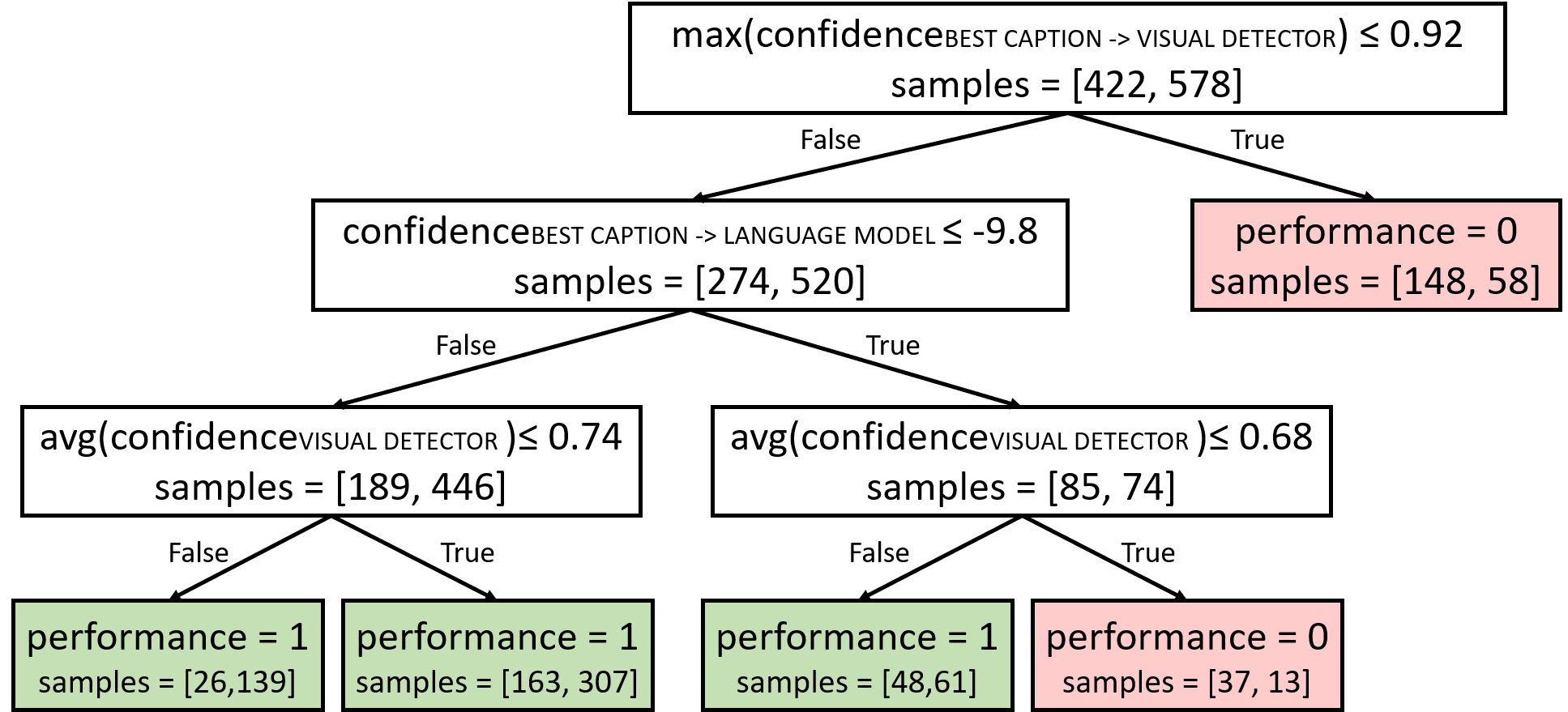}
      \caption{Single-model component view with system data.}
      \label{fig:component_nongt}
  \end{figure}

To illustrate more general failure situations, Figure~\ref{fig:component_nongt} displays a simplified generic-model tree from a component view with system data (component confidences). The tree reveals interesting condition rules to the system designer. For example, it conveys that the system is more likely to fail for instances where the terms mentioned in the final best caption have a low confidence from the Visual Detector ($\leq$ 0.92). 
  
%\noindent {\bfseries{Result 9:}} Individual decision trees from performance views uncover frequent conditions of failure. System designers can then use them as a debugging tool for identifying such conditions and based on their frequency decide whether it is necessary to take corrective actions.

\section{Related Work}
\noindent {\bfseries{Failure explanation.}} There has been increasing interest in the intelligibility of algorithmic decision making in the private and public sectors \cite{goodman2016eu}. The European Union's General Data Protection Regulation\footnote{https://www.eugdpr.org/} (GDPR) contains articles on the right of explanation for consequential decisions made in an automated manner. DARPA's project on Explainable AI \cite{gunning2017explainable} is aimed at stimulating research to create a suite of machine learning techniques that produce more explainable models, while maintaining a high level of performance. These interests are closely related to the challenges of transparency and interpretability in machine learning \cite{caruana2015intelligible,ribeiro2016should,ribeiro2016model,poursabzi2017manipulating,lakkaraju2017interpretable}. 

%In this work, we put a special focus on the problem of failure explanation as a step forward towards accountable machine learning. 
Previous work in error detection \cite{zhang2014predicting,bansal2014towards} and model explanation \cite{baehrens2010explain} focused on individual models. \mname widens the scope with analyses of failures for integrative systems that leverage multiple components. Related work on component-based systems have been aimed at finding weakest links in learning pipelines \cite{parikh2011finding} and troubleshooting by identifying the best components to fix \cite{DBLP:conf/aaai/NushiKHK17}. \mname is complementary to these approaches by providing rich, detailed explanations of error conditions.
Related work includes efforts on predicting the behavior of problem-solving systems \cite{horvitz2001bayesian,kautz2002dynamic} via learning jointly from evidence about attributes of input instances and about system operation. Other works have used decision trees \cite{chen2004failure}, support vector machines \cite{widodo2007support}, and Bayesian networks \cite{breese1996decision} for diagnosing (non-learning) machine failure.

\noindent {\bfseries{Human-AI systems.}} Human computation has traditionally been an integral part of artificial intelligence systems as a means of generating labeled training data \cite{deng2009imagenet,DBLP:conf/eccv/LinMBHPRDZ14}. Most recently, there is increasing interest in human computation as a framework for enabling hybrid intelligence systems \cite{kamar2012combining,kamar2016directions} as forms of fundamental human-machine collaboration. A recent study \cite{vaughan2017making} on use cases for crowdsourcing emphasizes potential and current benefits of seamlessly integrating human computation with machine learning. Related efforts include research on interactive machine learning \cite{amershi2014power,ware2001interactive} for building systems that interactively learn from their users. 

In the context of system performance analysis, human input has been leveraged for system evaluation \cite{steinfeld2007evaluation,shinsel2011mini} and testing \cite{groce2014you,attenberg2011beat}. Other relevant work on system explanation \cite{kulesza2015principles} introduces the concept of explanatory debugging, where the system explains to users the rationale behind predictions of a single classifier, and users correct erroneous outputs for personalizing the system. Our approach extends these efforts by explaining the behavior of component-based systems rather than single classifiers.

\section{Conclusion and Future Work}
We presented \mname, a hybrid human-machine approach to analyzing and explaining failure in component-based machine learning systems.
The methodology can provide machine learning practitioners with insights about performance via a set of views that reveal details about failures related to the system input and its internal execution. We demonstrated \mname with an application to an image captioning system. The results show the power of \mname as a new kind of lens on the performance of component-based AI systems that can reveal details about failures hidden in the aggregate statistics of traditional metrics. 

We see great opportunity ahead for employing the methods to inspect and refine the operation of inferential pipelines. Future directions with this work include the conceptualization of views that can jointly cluster different types of failures and system execution signals and extensions of the methods to provide insights about less interpretable, monolithic learning systems.

%We presented \mname, a hybrid human-machine method for analyzing and explaining failure in component-based machine learning systems.
%The approach provides machine learning practitioners with insightful performance views, which uncover hidden conditions of failure related to the system input and its internal execution. We applied \mname to a real-world image captioning system and the results revealed new facts about system performance and demonstrated the benefits of moving away from aggregate success metrics to multi-faceted detailed error analyses. 

%In the future, we plan to conceptualize views that can jointly cluster types of failures and system execution signals. Another important research direction is to study the applicability of \mname as a debugging tool for  monolithic and not yet interpretable end-to-end learning systems.
\bibliographystyle{aaai} 
\bibliography{document}
\flushend
\end{document}